\documentclass[conference]{IEEEtran}
\IEEEoverridecommandlockouts

\usepackage{cite}
\usepackage{amsmath,amssymb,amsfonts}
\usepackage{algorithmic}
\usepackage{graphicx}
\usepackage{textcomp}
\usepackage{xcolor}

\usepackage{threeparttable}
\usepackage{hhline}
\usepackage{pgfplots}
\usepackage{url}
\usepackage[nolist, printonlyused]{acronym}  
\usepackage{balance}

\usepackage{ifthen}
\usepackage{amssymb}
\usepackage{xcolor}
\newboolean{showcomments}
\setboolean{showcomments}{false}
\ifthenelse{\boolean{showcomments}}
{ \newcommand{\mynote}[3]{
     \fbox{\bfseries\sffamily\scriptsize#1}
        {\small$\blacktriangleright$\textsf{\emph{\color{#3}{#2}}}$\blacktriangleleft$}}}
{ \newcommand{\mynote}[3]{}}

\newcommand{\tl}{\raise.17ex\hbox{$\scriptstyle\mathtt{\sim}$}}

\def\BibTeX{{\rm B\kern-.05em{\sc i\kern-.025em b}\kern-.08em
    T\kern-.1667em\lower.7ex\hbox{E}\kern-.125emX}}

\makeatletter
\def\ps@IEEEtitlepagestyle{%
\def\@oddfoot{\mycopyrightnotice}%
\def\@evenfoot{}%
}
\def\mycopyrightnotice{%
{\footnotesize}
\gdef\mycopyrightnotice{}
}

\begin{document}

\title{Pruned Lightweight Encoders for Computer Vision}  

\author{\IEEEauthorblockN{Jakub Žádník, Markku Mäkitalo, Pekka Jääskeläinen}
\IEEEauthorblockA{\textit{Faculty of Information Technology and Communication Sciences}\\
\textit{Tampere University}, Finland \\
\texttt{\{jakub.zadnik, markku.makitalo, pekka.jaaskelainen\}@tuni.fi}}
}

\maketitle

\begin{abstract}
  Latency-critical computer vision systems, such as autonomous driving or drone control, require fast image or video compression when offloading neural network inference to a remote computer.
  To ensure low latency on a near-sensor edge device, we propose the use of lightweight encoders with constant bitrate and pruned encoding configurations, namely, ASTC and JPEG XS.
  Pruning introduces significant distortion which we show can be recovered by retraining the neural network with compressed data after decompression.
  Such an approach does not modify the network architecture or require coding format modifications.
  By retraining with compressed datasets, we reduced the classification accuracy and segmentation mean intersection over union (mIoU) degradation due to ASTC compression to 4.9--5.0 percentage points (pp) and 4.4--4.0 pp, respectively.
  With the same method, the mIoU lost due to JPEG XS compression at the main profile was restored to 2.7--2.3 pp.
  In terms of encoding speed, our ASTC encoder implementation is 2.3x faster than JPEG.
  Even though the JPEG XS reference encoder requires optimizations to reach low latency, we showed that disabling significance flag coding saves 22--23\% of encoding time at the cost of 0.4--0.3 mIoU after retraining.
\end{abstract}


\section{Introduction}
\label{sec:intro}

\begin{figure}[ht]
    \centering
    \includegraphics[width=\linewidth]{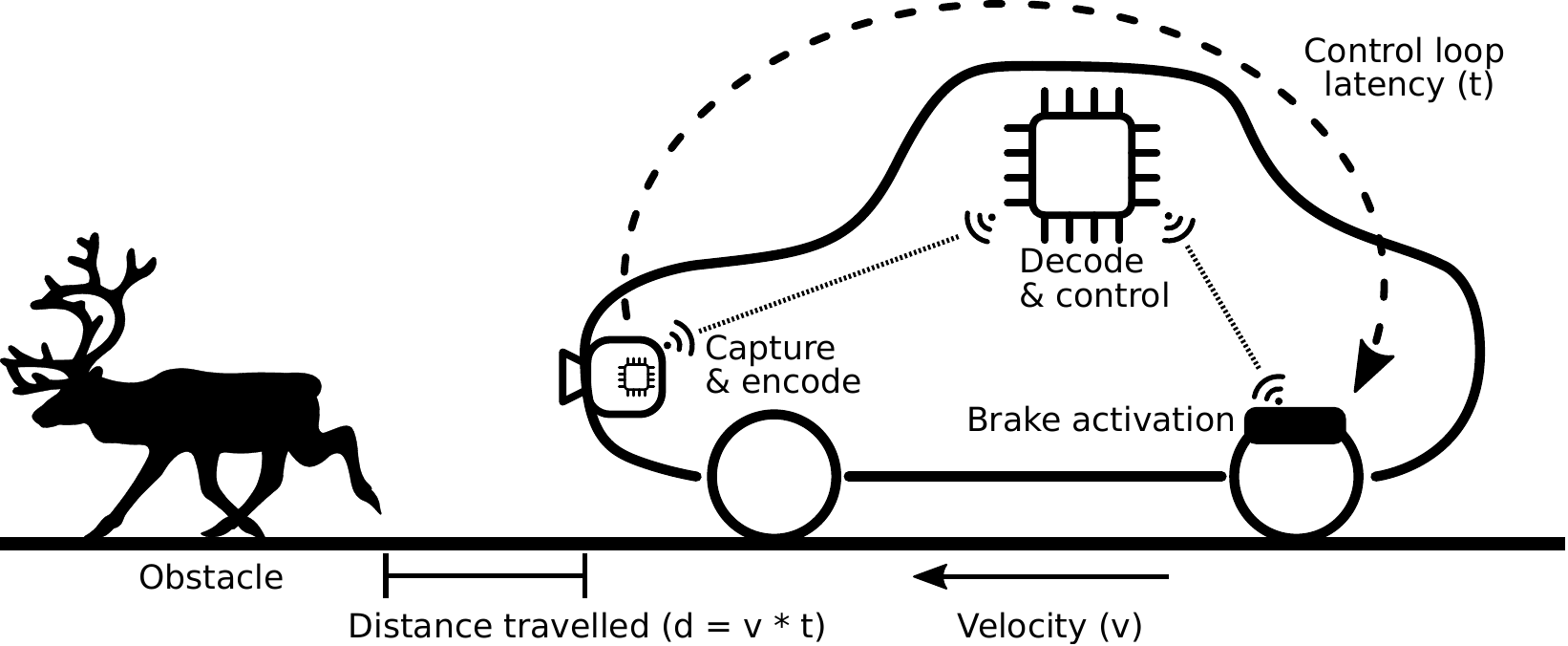}
    \caption{The reaction time of an autonomous vehicle control system determines whether it can avoid hitting an obstacle.}
    \label{fig:system}
  \end{figure}

  Automated systems that make fast decisions based on visual input, such as autonomous driving, drone control, or smart factories, rely on a very short response time to prevent damage or injury.
  Low-latency network transmission enabled by recent development in networking technologies, such as 5G, allows edge devices with low computing power to offload expensive \ac{DNN} inference of the vision task to a nearby server.
  Figure~\ref{fig:system} depicts a model scenario of an obstacle suddenly emerging in a trajectory of a self-driving car.
  Considering the car's speed of 100 km/h, if the end-to-end latency of the brake control system increased by 40 ms, for example, due to slow compression, the car would travel an additional 1.1 meters, potentially hitting the obstacle instead of stopping in front of it.


  Fast compression of source images is necessary to ensure low latencies over a transfer channel, and one way to decrease the latency is to reduce the codec complexity.
  Removing coding features, however, typically results in decreased vision performance at the same bitrate.
  Since \acp{DNN} have the ability to learn from the input data, it is possible to retrain them on the compressed dataset (after decompressing it) to overcome the coding efficiency lost by pruning the coding features.
  At the same time, pruning the existing codecs allows to reuse existing hardware support and retraining does not modify the neural network architecture since only the weights change.

  Several types of low-complexity codecs exist.
  Real-time texture compression can reach very high encoding speeds compared to other methods at the expense of rather low coding efficiency~\cite{waveren2007, holub2013, zadnik2022}.
  A ``mezzanine compression'' family of codecs is designed specifically to meet ultra-low latency requirements, with \acs{JPEG}~XS~\cite{descampe2021jpeg} as the newest standard in this family.
  The recently standardized \ac{HTJ2K}~\cite{htj2k} simplifies the otherwise complex \acs{JPEG}~2000~\cite{skodras2001jpeg} with the goal of 10x throughput improvement.
  However, it does not offer such a precise rate allocation as \acs{JPEG}~XS.
  Traditional hybrid video codecs, such as \ac{HEVC}~\cite{sullivan2012} or \ac{VVC}~\cite{bross2020} offer advanced coding features and great coding efficiency.
  However, the additional complexity of, for example, the inter or intra prediction and advanced entropy coding can be prohibitive in resource-constrained devices.
  \Ac{JPEG}~\cite{wallace1992} shares the core transform coding features with hybrid video codecs, but without the additional complexity. 
  At the same time, it can deliver sufficient quality for computer vision applications, as shown in the paper.

  In this paper, we explore the idea of pruning the encoding configurations to reduce the encoding time and latency and compensating the lost vision performance by retraining the vision model with the compressed dataset.
  As two case studies we chose reducing the configuration space of the otherwise very complex \ac{ASTC}~\cite{nystad2012} format and \ac{JPEG}~XS that was designed specifically as a lightweight, low-latency codec.
  Both codecs operate in a constant bitrate mode which is important for ensuring predictable latency.

  We evaluate the effect of \ac{ASTC} compression artifacts on the image classification accuracy of ShuffleNet~\cite{ma2018shufflenet} V2 and both \ac{ASTC} and \ac{JPEG}~XS on the semantic segmentation of LR-ASPP-MobileNetV3~\cite{howard2019}.
  We compare both to \ac{JPEG}, and in the case of the segmentation task also to \ac{HTJ2K} and \ac{JPEG}~2000.

  The contributions of this paper are:
  \begin{itemize}
      \item We propose a lightweight \ac{ASTC} encoder\footnote{\url{github.com/cpc/simple-texcomp}} that is approximately $2.3\times$ faster than \ac{JPEG} on a Samsung S10 smartphone.
      \item We study how pruning \ac{JPEG}~XS encoding configurations impacts latency and computer vision performance.
      \item We demonstrate that the quality vs. latency tradeoff can be alleviated by retraining the classification and segmentation models with the compressed datasets. 
  \end{itemize}

\section{Background and Related Work}

\subsection{Adaptive Scalable Texture Compression}

\ac{ASTC} is the newest and most flexible texture compression format adopted as an OpenGL extension by the Khronos Group.
Like other texture compression formats, it quantizes the input block's color space and represents its pixels as indices pointing at one of the quantized colors.
Compared to the older BCn formats, \ac{ASTC} supports many configuration options: scaling the input block size, partitioning, different color endpoint modes (CEM), endpoint and weight quantizations, dual-plane encoding, and \ac{BISE}.
An important property of texture compression is a fixed compression ratio and random access: The individual pixels are randomly addressable from the compressed representation without decompressing.

Modern \acp{GPU} have texture fetch units that can perform the decompression online during rendering with a negligible overhead which further enhances the low-latency potential of texture compression.
\subsection{JPEG XS}

\ac{JPEG}~XS is a wavelet-based mezzanine codec designed primarily for low complexity, low latency, high bandwidth, and high-quality video delivery.
The minimal coding unit of \ac{JPEG}~XS is one precinct whose size can range from less than one pixel line up to several lines of the image.

The \ac{JPEG}~XS rate allocation can predict the bitrate precisely, unlike \ac{HTJ2K} where a precise rate allocation would require a significant additional complexity~\cite{descampe2021jpeg}.
The latest version of the standard also supports direct Bayer data compression~\cite{richter2021bayer} which can be used to bypass the traditional image processing pipeline at the sensor side and thus save latency.

To the best of our knowledge, no publicly available JPEG XS encoder currently exists.
Therefore, we utilized the reference \ac{JPEG}~XS reference software, version 1.4.0.4 (ISO/IEC 21122-5:2020).
In the literature, \cite{itakura2020} developed a JPEG XS codec capable of running at 60 \ac{FPS} at 8K resolution on a 64-core AMD EPYC processor.


\subsection{Compression for Computer Vision}
\label{subs:cv}


Some previous works optimize the perceptual model of \ac{JPEG} for computer vision~\cite{xie2019, liu2018}, leading to significant quality improvements.
\cite{brummer2020} optimized the global JPEG XS encoding parameters (gains and priorities) to better capture the characteristics of a computer vision target.
Our approach of retraining the vision model with the compressed dataset is complementary to codec parameter optimizations.

\cite{zadnik2021} used retraining to recover object detection and semantic segmentation performance of BC1 and YCoCg-BC3~\cite{waveren2007} texture compression.
To the best of our knowledge, no prior work implements a minimal-subset \ac{ASTC} in the context of computer vision.

\cite{marie2022expert} proposed a modified loss function of a \ac{DNN} to achieve more efficient restoration of classification accuracy lost to compression artifacts.
They achieved a minor but consistent gain of up to 0.79 \ac{pp} validation accuracy compared to a simple retraining method used in this work.

The recent exploration of \ac{VCM} by \ac{MPEG} is an effort to develop a coding scheme with both machine and human perception in mind~\cite{duan2020}.
The current development is being built on top of \ac{VVC} which is a more complex format than what we target in this paper.
Furthermore, our use case considers only the computer vision performance without the human in the loop.
JPEG AI \cite{jpegai2021} also explores compression for both human and computer vision targets, but focuses on utilizing learning-based coding methods.

Yet another approach to adapting compression for computer vision is ``feature compression'' which encodes intermediate neural features~\cite{shao2020bottlenet}.
Feature compression, however, requires computing some of the convolutional layers on the encoding device which contrasts with our approach of decreasing the encoding complexity.

\section{Implementation of Pruned Codecs}

\subsection{ASTC}

Due to the \ac{ASTC} complexity, exhaustively searching for encoding parameters is not feasible in real time, and such, heuristics must be used to prune the configuration space.
In our work, we reduce the configuration space to only one configuration: 5-bit color endpoint and 2-bit and weight quantization with a weight grid of $8\times5$.
The selected configuration showed the lowest per-pixel distortion measured as \ac{PSNR} on a sample dataset from adjacent configurations without requiring \ac{BISE}.

Since the only way to scale the \ac{ASTC} bitrate is to modify the input block size, we implemented both $12\times12$ and $8\times8$ input block sizes, implying a \ac{CR} of 27:1 and 12:1 ($0.\overline{8}$ and 2.0 \ac{bpp}), respectively.

The encoding of a block starts by selecting the endpoints with a small inset similar to~\cite{waveren2007}.
Then, ``ideal weights'' are selected by orthogonally projecting the input pixels onto the line defined by the endpoints.
Lastly, the ``ideal weights'' are bilinearly downsampled to the $8\times5$ grid and quantized into two bits.


\subsection{JPEG XS}


The long encoding time of the reference encoder is caused by the rate allocation algorithm exhaustively computing the bit budget for each precinct at all quantization levels and using all possible coding methods.
We reduced the number of searched quantizations and coding methods to 13 and 5, respectively, without losing quality as the other combinations were unused in our tests.

To reduce the number of rate allocation passes further, we disabled the significance flag coding.
Significance flag coding detects a run of all-zero ``significance groups'' (groups of 8 adjacent coefficients) that can be encoded with a single flag and requires an additional rate allocation pass.
Disabling this method brings the number of utilized coding methods from 5 to 3 and removes the need for ``refresh'' passes,  significantly reducing the encoding time.
However, it also reduces the coding efficiency, which we try to recover by retraining with the compressed dataset.
We kept the coefficient prediction from a previous line, as disabling it would prevent the encoder meet the target bitrate.


\section{Experimental Setup}

\subsection{Implementation Details}
\label{subs:impl}


The \ac{ASTC} encoder uses ARM NEON intrinsics to vectorize the most significant loops using 8-bit fixed-point representation.
For a fair runtime comparison with \ac{JPEG}, we chose the \ac{SIMD}-optimized \texttt{libjpeg-turbo} library\footnote{\url{libjpeg-turbo.org} (version 2.1.1)} and developed a wrapper encoder application around the library.
The \ac{JPEG} coding parameters were chosen to match the defaults of the \texttt{cjpeg} command-line utility: YCbCr color space with 4:2:0 subsampling and no restart intervals.
The quality parameter (Q) 45 was selected so that the bitrate of a random sample of 10000 ImageNet images after encoding is the highest possible at or below the rate of \ac{ASTC} $12\times12$.
Both \ac{ASTC} and \ac{JPEG} were evaluated on a single core (A76) of a Samsung S10 smartphone.



The \ac{JPEG}~XS encoder was compiled only for the x86 instruction set and evaluated on a single thread of Intel i7-8650U laptop CPU at a base frequency of 2.1 GHz with disabled frequency scaling.
For runtime comparison we chose two open source encoders: \texttt{grok}\footnote{\url{github.com/GrokImageCompression/grok} (version 9.7.7)} for \ac{JPEG}~2000 and \texttt{OpenJPH}\footnote{\url{github.com/aous72/OpenJPH} (version 0.9.0)} for \ac{HTJ2K}, both using irreversible \ac{DWT}.





\subsection{Vision Tasks}

We evaluated the image classification accuracy of ShuffleNet V2\footnote{\url{pytorch.org/hub/pytorch_vision_shufflenet_v2}} in $0.5\times$ and $1.0\times$ sizes trained on the ImageNet dataset~\cite{deng2009imagenet} with training hyperparameters derived from \cite{zhang2018shufflenet} and \cite{ma2018shufflenet}.
We also evaluated a semantic segmentation task with LR-ASPP-MobileNetV3\footnote{\url{github.com/ekzhang/fastseg} (commit 91238cd)} in both large and small versions trained on the Cityscapes dataset~\cite{cordts2016}.
The Cityscapes images for training were cropped to $\sqrt{2}$ of the original size in each dimension to avoid running out of \ac{GPU} memory.
Both networks were retrained with the dataset compressed with \ac{ASTC} to recover the vision performance lost by compression artifacts.
Unfortunately, the \ac{JPEG}~XS encoder was not able to encode some of the ImageNet images at the bitrate of $0.\overline{8}$ \ac{bpp}.
Therefore, we evaluated only the segmentation task with this codec.

All encoders mentioned in the previous subsection were used for quality evaluations, along with \texttt{astcenc}\footnote{\url{github.com/ARM-software/astc-encoder} (version 3.7)} at the fastest profile for quality evaluations on the Cityscapes dataset.

\ac{JPEG}-compressed images were used only for retraining the ShuffleNet V2 network.
\ac{JPEG}, \ac{JPEG}~2000, and \ac{HTJ2K} reach segmentation \ac{mIoU} within 1.5\% below the uncompressed result, and retraining is expected to bring the results on par with the uncompressed results, therefore, we did not retrain with these codecs.

\section{Results}

\subsection{Quality}

\paragraph*{Image Classification}



Table~\ref{tab:quality_shufflenet} summarizes the highest achieved classification accuracies of ShuffleNet V2 under different conditions.
When the compressed data is used as an input to the network trained on uncompressed data (the ``orig.'' column), the \ac{ASTC} compression degrades the accuracy by more than 15 \ac{pp}, while the difference caused by \ac{JPEG} compression is only 1.3 and 0.6 \ac{pp}.
However, when retrained with the compressed dataset (the ``retr.'' column), the accuracy decrease for \ac{ASTC} with $12\times12$ block size is only 4.9 and 5.0 \ac{pp} for the $0.5\times$ and $1.0\times$ network sizes, respectively.
The \ac{ASTC} $8\times8$ achieves higher quality than $12\times12$: only 2.3--1.8 \ac{pp} accuracy decrease compared to the uncompressed result.
Retraining with the \ac{JPEG}-compressed dataset brings a 0.2 \ac{pp} increase in the classification accuracy of the smaller network.
The results show that retraining the larger network does not improve the already high accuracy for \ac{JPEG}.

\begin{table}[htbp]
  \caption{Validation top-1 accuracy of ShuffleNet V2 on ImageNet validation set with JPEG and the proposed ASTC compression with and without retraining.}
  \label{tab:quality_shufflenet}
  \centering
  \begin{threeparttable}
      \begin{tabular}{|l|r|rr|rr|}
        \hline
            & & \multicolumn{2}{c|}{ 0.5x } & \multicolumn{2}{c|}{ 1.0x } \\
        compression & bpp & orig. & retr. & orig. & retr. \\
        \hhline{|=|=|==|==|}
        uncompressed &     24.0 & \multicolumn{2}{c|}{54.4\%} & \multicolumn{2}{c|}{64.3\%} \\
        ASTC 12x12   &     0.89 & -16.8 & -4.9 & -15.1 & -5.0 \\
        JPEG Q45     & \tl 0.89 &  -1.3 & -1.1 &  -0.6 & -0.7 \\
        ASTC 8x8     &     2.00 &  -6.4 & -2.3 &  -6.6 & -1.8 \\
        \hline
      \end{tabular}
  \end{threeparttable}
\end{table}

\paragraph*{Semantic Segmentation}

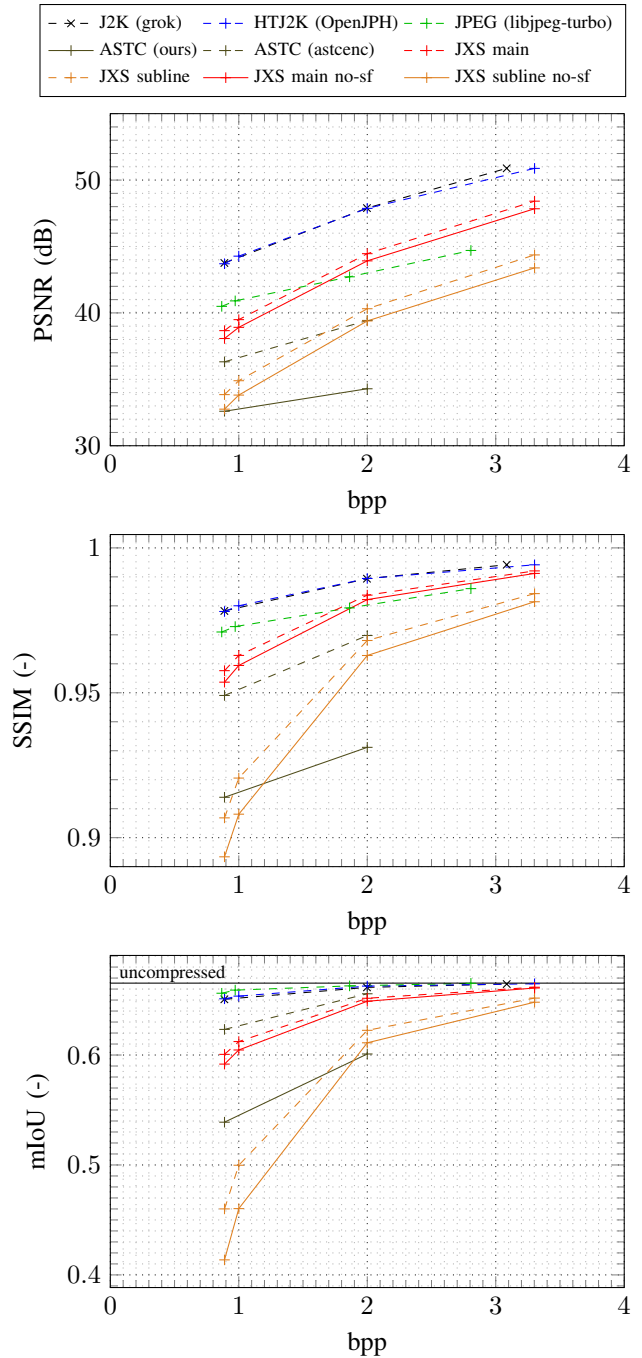
\begin{figure}[t]
  \centering
  \begin{tikzpicture}
    \tikzstyle{dataline} = [color=black, dashed, mark options=solid, mark=+]
    \begin{axis}
      [
        yshift           = -11cm,
        width=0.95\linewidth,
        height=6cm,
        xlabel           = {bpp},
        xlabel near ticks,
        ylabel           = {PSNR (dB)},
        ylabel near ticks,
        grid             = both,
        major grid style = {dotted,black!90},
        minor grid style = {dotted,gray!60},
        legend columns    = 3,
        legend style      = { at = {(0.433, 1.05)}, anchor = south, font = \scriptsize },
        legend cell align = { left },
        xmin = 0,
        xmax = 4,
        ymin = 30,
        ymax = 55,
        ytick distance = 10,
        minor x tick num = 9,
        minor y tick num = 9,
      ]

      \addplot [dataline, black, mark=x]
        table[x=bpp_mean, y=psnr_mean, col sep=comma] {data/generated/grok_jp2_irev.csv};
      \addplot [dataline, blue]
        table[x=bpp_mean, y=psnr_mean, col sep=comma] {data/generated/openjph.csv};
      \addplot [dataline, green!75!black]
        table[x=bpp_mean, y=psnr_mean, col sep=comma] {data/generated/turbojpeg.csv};
      \addplot [dataline, solid, black!40!purple!70!green]
        table[x=bpp_mean, y=psnr_mean, col sep=comma] {data/generated/astc_u8.csv};
      \addplot [dataline, black!40!purple!70!green]
        table[x=bpp_mean, y=psnr_mean, col sep=comma] {data/generated/astcenc_fastest.csv};
      \addplot [dataline, red]
        table[x=bpp_mean, y=psnr_mean, col sep=comma] {data/generated/jxs_p3.csv};
      \addplot [dataline, orange!50!brown]
        table[x=bpp_mean, y=psnr_mean, col sep=comma] {data/generated/jxs_p5.csv};
      \addplot [dataline, solid, red]
        table[x=bpp_mean, y=psnr_mean, col sep=comma] {data/generated/jxs_nosf_p3.csv};
      \addplot [dataline, solid, orange!50!brown]
        table[x=bpp_mean, y=psnr_mean, col sep=comma] {data/generated/jxs_nosf_p5.csv};

      \legend{ J2K (grok), HTJ2K (OpenJPH), JPEG (libjpeg-turbo), ASTC (ours),
        ASTC (astcenc), JXS main, JXS subline, JXS main no-sf, JXS subline no-sf }

    \end{axis}
    \begin{axis}
      [
        yshift           = -16.6cm,
        width=0.95\linewidth,
        height=6cm,
        xlabel           = {bpp},
        xlabel near ticks,
        ylabel           = {SSIM (-)},
        ylabel near ticks,
        grid             = both,
        major grid style = {dotted,black!90},
        minor grid style = {dotted,gray!60},
        xmin = 0,
        xmax = 4,
        ymin = 0.89,
        minor x tick num = 9,
        minor y tick num = 4,
      ]

      \addplot [dataline, black, mark=x]
        table[x=bpp_mean, y=ssim_mean, col sep=comma] {data/generated/grok_jp2_irev.csv};
      \addplot [dataline, blue]
        table[x=bpp_mean, y=ssim_mean, col sep=comma] {data/generated/openjph.csv};
      \addplot [dataline, green!75!black]
        table[x=bpp_mean, y=ssim_mean, col sep=comma] {data/generated/turbojpeg.csv};
      \addplot [dataline, solid, black!40!purple!70!green]
        table[x=bpp_mean, y=ssim_mean, col sep=comma] {data/generated/astc_u8.csv};
      \addplot [dataline, black!40!purple!70!green]
        table[x=bpp_mean, y=ssim_mean, col sep=comma] {data/generated/astcenc_fastest.csv};
      \addplot [dataline, red]
        table[x=bpp_mean, y=ssim_mean, col sep=comma] {data/generated/jxs_p3.csv};
      \addplot [dataline, orange!50!brown]
        table[x=bpp_mean, y=ssim_mean, col sep=comma] {data/generated/jxs_p5.csv};
      \addplot [dataline, solid, red]
        table[x=bpp_mean, y=ssim_mean, col sep=comma] {data/generated/jxs_nosf_p3.csv};
      \addplot [dataline, solid, orange!50!brown]
        table[x=bpp_mean, y=ssim_mean, col sep=comma] {data/generated/jxs_nosf_p5.csv};

    \end{axis}
    \begin{axis}
      [
        yshift           = -22.2cm,
        width=0.95\linewidth,
        height=6cm,
        xlabel           = {bpp},
        xlabel near ticks,
        ylabel           = {mIoU (-)},
        ylabel near ticks,
        grid              = both,
        major grid style  = {dotted,black!90},
        minor grid style  = {dotted,gray!60},
        xmin              = 0,
        xmax              = 4,
        minor x tick num  = 9,
        minor y tick num  = 9,
      ]

      \addplot [dataline, black, mark=x]
        table[x=bpp_mean, y=fastseg_large_miou, col sep=comma] {data/generated/grok_jp2_irev.csv};
      \addplot [dataline, blue]
        table[x=bpp_mean, y=fastseg_large_miou, col sep=comma] {data/generated/openjph.csv};
      \addplot [dataline, green!75!black]
        table[x=bpp_mean, y=fastseg_large_miou, col sep=comma] {data/generated/turbojpeg.csv};
      \addplot [dataline, solid, black!40!purple!70!green]
        table[x=bpp_mean, y=fastseg_large_miou, col sep=comma] {data/generated/astc_u8.csv};
      \addplot [dataline, black!40!purple!70!green]
        table[x=bpp_mean, y=fastseg_large_miou, col sep=comma] {data/generated/astcenc_fastest.csv};
      \addplot [dataline, red]
        table[x=bpp_mean, y=fastseg_large_miou, col sep=comma] {data/generated/jxs_p3.csv};
      \addplot [dataline, orange!50!brown]
        table[x=bpp_mean, y=fastseg_large_miou, col sep=comma] {data/generated/jxs_p5.csv};
      \addplot [dataline, solid, red]
        table[x=bpp_mean, y=fastseg_large_miou, col sep=comma] {data/generated/jxs_nosf_p3.csv};
      \addplot [dataline, solid, orange!50!brown]
        table[x=bpp_mean, y=fastseg_large_miou, col sep=comma] {data/generated/jxs_nosf_p5.csv};
      \addplot[mark=none, black, domain=0:4] {0.66539};

      \draw (0, 260) node [right] {\scriptsize uncompressed};

    \end{axis}
  \end{tikzpicture}
  \caption{Mean intersection over union (mIoU) of a FastSeg large network (bottom), SSIM (middle) and PSNR (top) of a Cityscapes validation set compressed with different methods.}
  \label{fig:fastseg_eval}
\end{figure}

Figure~\ref{fig:fastseg_eval} compares rate-distortion curves of multiple encoders according to three metrics: \ac{PSNR}, \ac{SSIM}, and validation \ac{mIoU} of LR-ASPP-MobileNetV3 (large vesion) trained on an uncompressed dataset.
The small version of the model shows similar relations between the \ac{mIoU} curves to the large version, therefore, we omitted it for brevity.
The plots show that despite significant \ac{PSNR} and \ac{SSIM} differences between \ac{JPEG}~2000, \ac{JPEG}, and \ac{HTJ2K}, the \ac{mIoU} difference is relatively small.
Disabling significance flag coding of \ac{JPEG}~XS (denoted as ``no-sf'') shows a consistent decrease of all metrics in both main and subline profiles.
Similarly, when compared to a full-featured \texttt{astcenc} encoder at the fastest preset, our pruned implementation achieves significantly lower quality.
Both \ac{SSIM} and \ac{mIoU} metrics decrease rapidly with \ac{JPEG}~XS at lower bitrates ($0.\overline{8}$ and 1.0 \ac{bpp}), especially the subline profile.


To recover the large \ac{mIoU} degradation of \ac{ASTC} and \ac{JPEG}~XS at low bitrates, we retrained the network with the compressed datasets.
Table~\ref{tab:quality_mobilenet} summarizes the \ac{mIoU} improvements of retraining LR-ASPP-MobileNetV3 in both small and large variants.
For \ac{ASTC} $12\times12$, retraining brought an improvement of 1.1 and 8.6 \ac{pp} for the small and large networks, respectively.
Retraining \ac{JPEG}~XS in the main profile resulted in \ac{mIoU} around 2.3--2.6 \ac{pp} lower than the uncompressed result.
The subline profile of \ac{JPEG}~XS shows a sharp decline in the vision performance without retraining.
Retraining allows recovering most of the quality back.
However, the results still do not reach the quality of \ac{ASTC} $12\times12$.


\begin{table}[htbp]
  \caption{Mean intersection over union (mIoU) of LR-ASPP-MobileNetV3 with Cityscapes validation set and the JPEG XS compression.}
  \label{tab:quality_mobilenet}
  \centering
  \begin{threeparttable}
      \begin{tabular}{|l|r|rr|rr|}
        \hline
            & & \multicolumn{2}{c|}{ small } & \multicolumn{2}{c|}{ large } \\
        compression              & bpp   & orig. & retr. & orig. & retr. \\
        \hhline{|=|=|==|==|}
        uncompressed             & 24.0  & \multicolumn{2}{c|}{61.2\%} & \multicolumn{2}{c|}{66.5\%} \\
        JPEG XS (main, sf)       & 0.89 & -4.9  & -2.3  & -6.5  & -2.6  \\
        JPEG XS (main, no-sf)    & 0.89 & -5.8  & -2.7  & -7.4  & -2.3  \\
        JPEG XS (subline, sf)    & 0.89 & -14.4 & -5.4  & -20.5 & -5.3  \\
        JPEG XS (subline, no-sf) & 0.89 & -17.3 & -6.7  & -25.2 & -6.8  \\
        ASTC 12x12               & 0.89 & -5.5  & -4.4  & -12.6 & -4.0  \\
        ASTC 8x8                 & 2.00 & -3.5  & -2.8  & -6.4  & -1.7  \\
        \hline
      \end{tabular}
  \end{threeparttable}
\end{table}

\begin{figure}
    \centering
    \includegraphics[width=\linewidth]{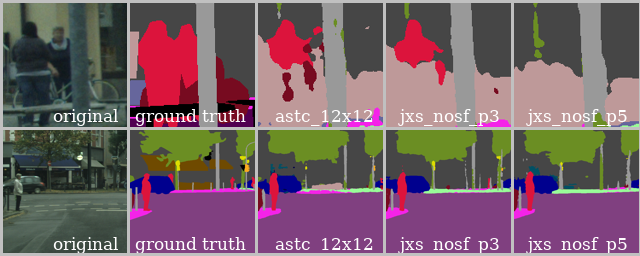}
    \caption{Visual comparison of LR-ASPP-MobileNetV3 (small version) segmentation of two Cityscapes images. From left to right: Original image (brightened), ground truth, segmentation of the model trained by datasets compressed by the ASTC ($12\times12$), pruned JPEG XS (main profile) and pruned JPEG XS (subline profile) encoders.}
    \label{fig:images}
\end{figure}

Figure~\ref{fig:images} shows segmentations of two challenging scenes after the retrained model inference using compressed images to illustrate the effect of compression on the segmentation result.
The first image shows shape deformations caused by the \ac{ASTC} and \ac{JPEG}~XS main profile (``p3'') while the network trained with \ac{JPEG}~XS subline profile (``p5'') fails to detect the people at all.
In the second image, all cases detect the person in the foreground but fail to detect some, or all, the people in the distance.
It should be emphasized that the MobileNet networks were trained with images containing approximately half of the pixels of the full-resolution images due to \ac{GPU} memory limitations.
Therefore, the examples do not correspond to the best predictions achievable with these networks.

\subsection{Runtime}

\paragraph*{ASTC Encoder}

Table~\ref{tab:runtime_arm} compares the average encoding time of a Cityscapes image (resolution $2048\times1024$) of our \ac{ASTC} encoder with the block size $12\times12$ and $8\times8$ to \ac{JPEG} with the quality parameters Q 85 and 96.
The Q parameters were determined by the same procedure as in Subsection~\ref{subs:impl} to ensure approximately the same bitrate as \ac{ASTC} on a random subset of 100 images.
For the $12\times12$ block size, the images were padded to a resolution divisible by a block size of 12 before \ac{ASTC} encoding.

The results show our simple \ac{ASTC} $12\times12$ encoder is approximately $2.3\times$ faster than the \ac{JPEG} encoder based on \texttt{libjpeg-turbo}.
The \ac{JPEG} decoding was slightly slower than the encoding.
While we did not conduct \ac{ASTC} decoding measurements, in~\cite{zadnik2021}, we measured BC1 and YCoCg-BC3 decoding time of an 8K frame as less than 1 ms on a desktop \ac{GPU}.
\ac{ASTC} decoding is more complicated, but the decoding overhead is still expected to be close to negligible in comparison to the encoding.

\begin{table}[htbp]
  \caption{Encoding time of the proposed \ac{ASTC} encoder and \texttt{libjpeg-turbo} encoder and decoder (ARM A76 single-core).}
  \label{tab:runtime_arm}
  \centering
  \begin{threeparttable}
      \begin{tabular}{|l|rr|}
        \hline
                              & \multicolumn{2}{|c|}{bpp}  \\
                              & \tl0.89 & \tl2.0 \\
        \hhline{|=|==|}
        ASTC                  & 5.8     & 7.0    \\
        JPEG (enc, libjpeg-turbo) & 13.3    & 16.7   \\
        JPEG (dec, libjpeg-turbo) & 13.6    & 21.5   \\
        \hline
      \end{tabular}
  \end{threeparttable}
\end{table}


For comparison, we also measured the encoding and decoding time of \ac{JPEG} at quality parameters 0 and 100 as 11 and 22 ms, and 7 and 32 ms, respectively.
These numbers establish the encoding speed bounds of this format.

On a single core of Intel i7-8650U laptop \ac{CPU}, the AVX2-optimized \texttt{astcenc} encoder compressed one Cityscapes image at approximately 44 and 61 ms (block sizes $12\times12$ and $8\times8$, respectively) at the fastest preset, suggesting that both \ac{JPEG} and our pruned \ac{ASTC} encoders are faster than a traditional \ac{ASTC} encoder even with the latter evaluated on a more powerful \ac{CPU}.

\paragraph*{JPEG XS Encoder}


Table~\ref{tab:runtime_x86} compares \ac{JPEG}~XS with three encoders: \ac{JPEG}, \ac{JPEG}~2000, and \ac{HTJ2K}.
The \ac{HTJ2K} quantization was determined by a similar procedure as in Subsection~\ref{subs:impl}.
The results show that by disabling the significance flag coding, the encoding time of one Cityscapes frame improved by 22--23\%, and is only about 9--20\% slower than \ac{JPEG}~2000.
\ac{HTJ2K} encoding by OpenJPH was 3.8--$6.2\times$ faster than \ac{JPEG}~XS without the significance coding flags.
\ac{JPEG} by \texttt{libjpeg-turbo} brought this difference further by almost another order of magnitude.
Kakadu \ac{HTJ2K} is not publicly available, therefore, we used results published in~\cite{taubman2019high} and extrapolated them to our setup.
More specifically, we scaled the result by the number of pixels from 4K to $2048\times1024$, multiplied by 4 since the original result was obtained on a 4-core machine, and finally scaled to our frequency of 2.1 GHz from the original 3.4 GHz.
Based on this rough estimation, it seems likely the \ac{HTJ2K} encoder is capable of reaching encoding throughput close to \ac{JPEG}.

\begin{table}[htbp]
  \caption{Encoding times of JPEG XS, JPEG, JPEG 2000, and HTJ2K encoders on a single core of i7-8650U CPU}
  \label{tab:runtime_x86}
  \centering
  \begin{threeparttable}
      \begin{tabular}{|l|rrr|}
        \hline
                                              & \multicolumn{3}{|c|}{bpp}   \\
                                              & \tl0.89 & \tl2.0        & \tl3.3 \\
        \hhline{|=|===|}
        JPEG XS (main, sf)                    & 625     & 654           & 683    \\
        JPEG XS (main, no-sf)                 & 477     & 503           & 531    \\
        JPEG (libjpeg-turbo)                  & 11.3    & 13.5          & 16.0   \\ 
        JPEG 2000 (grok)                      & 437     & 439           & 441    \\ 
        HTJ2K (OpenJPH)                       & 73.7    & 105           & 138    \\
        HTJ2K (Kakadu~\cite{taubman2019high}) & -       & 14.4\tnote{*} & -      \\
        \hline
      \end{tabular}
      \begin{tablenotes}[para]
        \item[*] extrapolated from the result in the publication
      \end{tablenotes}
  \end{threeparttable}
\end{table}

Table~\ref{tab:latency} shows \ac{JPEG}~XS encoding time and latency at two different bitrates and three profiles: high, main, and subline.
The ``precinct'' column denotes the number of lines that form one presinct.
The ``enc'' column shows the total frame encoding time and the ``latency'' column shows the time until the first precinct is done encoding and thus represents the minimum theoretical achievable latency.
While the overall frame encoding time does not differ dramatically between presets, the precinct size has a major impact on latency: The precinct of the high profile consisting of three lines shows a latency of 32--33\% of the total encoding time, while the latency of a half-line precinct of the subline profile is three times smaller portion of the encoding time.
Thus, even without reaching a high throughput, it is possible to achieve latency almost an order of magnitude lower than the encoding time.
It should also be noted that the latency includes the wavelet transform over the whole frame and can be further reduced by pipelining it with the rest of the computation.

\begin{table}[htbp]
  \caption{Latency and throughput comparison of the pruned JPEG XS reference encoder without significance flag coding.} 
  \label{tab:latency}
  \centering
  \begin{threeparttable}
      \begin{tabular}{|lll|rrr|}
        \hline
                  & precinct & bpp  & enc  & \multicolumn{2}{c|}{latency}      \\
        profile   & (lines)  &      & (ms) & (ms) & (\%) \\
        \hhline{|===|===|}
          high    & 3        & 0.89 & 502  & 167  & 33\%  \\
          high    & 3        & 2.0  & 528  & 166  & 31\%  \\
          main    & 2        & 0.89 & 477  & 136  & 29\%  \\
          main    & 2        & 2.0  & 504  & 137  & 27\%  \\
          subline & 0.5      & 0.89 & 443  & 53   & 12\%  \\
          subline & 0.5      & 2.0  & 469  & 53   & 11\%  \\
        \hline
      \end{tabular}
  \end{threeparttable}
\end{table}

\section{Discussion}
\label{sec:discussion}




Retraining with the compressed dataset showed the largest improvements when the vision performance without retraining was very low, such as the classification with the \ac{ASTC} $12\times12$ and segmentation with the subline \ac{JPEG}~XS encoders.
On the segmentation task, the overall quality decrease without retraining was smaller, because the images are less noisy and less prone to compression artifacts.
The vision performance could be further enhanced by improving the retraining process and optimizing encoding parameters for computer vision using one of the methods introduced in Subsection~\ref{subs:cv}.

The lightweight \ac{ASTC} encoder achieves a higher encoding speed than \ac{JPEG}, making it the fastest encoder evaluated.
On the other hand, the pruned \ac{JPEG}~XS reference encoder does not achieve sufficient runtime performance.
However, its low complexity and results from literature~\cite{itakura2020} suggest a fast implementation is possible.
The rate allocation can be further optimized by, for example, replacing the exhaustive search with a binary search, in combination with other heuristics. 
The second most expensive operation in the reference encoder is the wavelet transform which we did not modify.
In the reference encoder, the wavelet transform is performed over the whole frame before the coding of individual precincts.
However, it is possible to interleave the wavelet transform with the precinct coding as the latency of the wavelet transform ranges from a few pixels to 6 lines~\cite{descampe2021jpeg}.

To put the results into a practical perspective, let us consider a scenario of compressing a Cityscapes image and sending it over a 500 Mbit/s commercially available 5G network and an embedded transceiver capable of 2 Mbit/s.
Assuming a 1 ms latency budget for encoding and network transfer, the \ac{ASTC} at the bitrate of $0.\overline{8}$ would require a latency of 145 and 3.2 lines, respectively, assuming the encoding speed of 5.8 ms/frame.
While the first case allows partitioning the image into larger chunks, in the second case, as shown in Figure~\ref{fig:fastseg_eval}, lowering the latency of \ac{JPEG}~XS comes at a significant quality loss and thus necessitates the compensation by retraining.
\section{Conclusion}

We explored decreasing an image encoder complexity to achieve lower latency.
Namely, we evaluated a lightweight implementation of an \ac{ASTC} encoder and a pruned version of a \ac{JPEG}~XS reference encoder.

The \ac{ASTC} encoder outperforms \ac{JPEG} in terms of encoding speed by approximately $2.3\times$ at the same bitrate.
When retrained with the dataset compressed with ASTC at the lowest bitrate of $0.\overline{8}$, the classification accuracy was about 5 \ac{pp}, and the segmentation \ac{mIoU} 4.4--4.0 \ac{pp} lower than the output of the networks trained and evaluated without any compression.

The pruned \ac{JPEG}~XS reference encoder is not nearly as fast as \ac{ASTC} and needs more optimizations to be usable for real-time tasks.
Nevertheless, we show that disabling significance flag coding decreases the number of required rate allocation passes, and boosts the encoding speed by 22--23\% at the cost of only 0.4--0.3 \ac{pp} of segmentation \ac{mIoU} after retraining.


\ac{HTJ2K} and \ac{JPEG} outperform both the tested codecs in terms of vision performance.
However, \ac{ASTC} still holds the advantage of the fastest coding speed, while \ac{JPEG}~XS, if sufficiently optimized, is suitable for applications requiring ultra-low latencies.
To improve the quality, it is possible to apply computer vision--specific encoding parameter optimizations or improve the retraining process.

\section*{Acknowledgment}

The work was financially supported by the Tampere University ITC Graduate School. It was also supported by European Union’s Horizon 2020 research and innovation program under Grant Agreement No 871738 (CPSoSaware) and in part by the Academy of Finland under Grant 325530.





\balance

\bibliographystyle{IEEEtran}
\bibliography{bibliography}

\begin{acronym}
  \acro{JPEG}{joint photographic experts group}
  \acro{PSNR}{peak signal-to-noise ratio}
  \acro{ASTC}{adaptive scalable texture compression}
  \acro{ASIC}{application-specific integrated circuit}
  \acro{CPU}{central processing unit}
  \acro{GPU}{graphics processing unit}
  \acro{HEVC}{high efficiency video coding}
    \acroindefinite{HEVC}{an}{a}
  \acro{AVC}{advanced video coding}
  \acro{bpp}{bits per pixel}
  \acro{FPS}{frames per second}
  \acro{mIoU}{mean intersection over union}
  \acro{SIMD}{single instruction multiple data}
  \acro{QP}{quantization parameter}
  \acro{NN}{neural network}
  \acro{DNN}{deep neural network}
  \acro{VCM}{video coding for machines}
  \acro{VVC}{versatile video coding}
  \acro{pp}{percentage points}
  \acro{CEM}{color endpoint mode}
  \acro{BISE}{bounded integer sequence encoding}
  \acro{DSP}{digital signal processing}
  \acro{ROI}{region of interest}
  \acro{CR}{compression ratio}
  \acro{HTJ2K}{high throughput JPEG 2000}
  \acro{DWT}{discrete wavelet transform}
  \acro{EBCOT}{embedded block coding with optimized truncation}
  \acro{FBCOT}{fast block coding with optimized truncation}
  \acro{SSIM}{structural similarity index}
  \acro{MPEG}{moving picture experts group}
\end{acronym}

\end{document}